\pdfoutput=1 
\documentclass[sigconf]{acmart}

\usepackage{multicol}

\usepackage{subfigure}
\usepackage{multirow}
\usepackage{array}
\usepackage{enumitem}
\usepackage{scalerel}
\usepackage{amsmath}
\usepackage{graphicx}
\usepackage{amsthm}

\usepackage{wrapfig}
\usepackage{xcolor}

\definecolor{myred}{RGB}{255, 40, 0}
\definecolor{mygreen}{RGB}{0, 168, 107}

\newcolumntype{L}[1]{>{\raggedright\let\newline\\\arraybackslash\hspace{0pt}}m{#1}}
\newcolumntype{C}[1]{>{\centering\let\newline\\\arraybackslash\hspace{0pt}}m{#1}}
\newcolumntype{R}[1]{>{\raggedleft\let\newline\\\arraybackslash\hspace{0pt}}m{#1}}

\AtBeginDocument{%
	\providecommand\BibTeX{{%
			\normalfont B\kern-0.5em{\scshape i\kern-0.25em b}\kern-0.8em\TeX}}}

\copyrightyear{2021}
\acmYear{2021}
\setcopyright{acmlicensed}
\acmConference[KDD '21] {Proceedings of the 27th ACM SIGKDD Conference on Knowledge Discovery and Data Mining}{August 14--18, 2021}{Virtual Event, Singapore.}
\acmBooktitle{Proceedings of the 27th ACM SIGKDD Conference on Knowledge Discovery and Data Mining (KDD '21), August 14--18, 2021, Virtual Event, Singapore}
\acmPrice{15.00}
\acmISBN{978-1-4503-8332-5/21/08}
\acmDOI{10.1145/3447548.3467436}

\begin{document}
\fancyhead{}
	\title{Mitigating Performance Saturation in Neural Marked Point Processes: Architectures and Loss Functions}
	
	\author{Tianbo Li}
	\authornote{Both authors contributed equally to this research.} \authornote{Corresponding author.}
	\authornote{This work  was done when he was  a student at Nanyang Technological University, Singapore.}
	\affiliation{%
		\institution{Sea AI Lab}
		\streetaddress{1 Fusionopolis Place, 138522}
		\country{Singapore}}
	\email{litb@sea.com}
	
	\author{Tianze Luo}
	\authornotemark[1]
	\affiliation{%
		\institution{Nanyang Technological University}
		\streetaddress{50 Nanyang Ave, 639798}
		\country{Singapore}}
	\email{tianze001@e.ntu.edu.sg}
	
	\author{Yiping Ke}
	\affiliation{%
		\institution{Nanyang Technological University}
		\streetaddress{50 Nanyang Ave, 639798}
		\country{Singapore}}
	\email{ypke@ntu.edu.sg}
	
	\author{Sinno Jialin Pan}
	\affiliation{%
		\institution{Nanyang Technological University}
		\streetaddress{50 Nanyang Ave, 639798}
		\country{Singapore}}
	\email{sinnopan@ntu.edu.sg}

	\renewcommand{\shortauthors}{Anonymous, et al.}
	
	\begin{abstract}
		Attributed event sequences are commonly encountered in practice. A recent research line focuses on incorporating neural networks with the statistical model---marked point processes, which is the conventional tool for dealing with attributed event sequences. Neural marked point processes possess  good interpretability of  probabilistic models as well as the representational power of neural networks. However, we find that performance of neural marked point processes is not always increasing as the network architecture becomes more complicated and larger, which is what we call the \textit{performance saturation} phenomenon. This is due to the fact that the generalization error of neural marked point processes is determined by both the network representational ability and the model specification at the same time. Therefore we can draw two major conclusions: first, simple network structures can perform no worse than complicated ones for some cases;  second, using a proper probabilistic assumption is as equally, if not more, important as improving the complexity of the network.	Based on this observation, we propose a simple graph-based network structure called GCHP, which utilizes only graph convolutional layers, thus it can be easily accelerated by the parallel mechanism.  We directly consider the distribution of interarrival times instead of imposing a specific assumption on the conditional intensity function, and propose to use a likelihood ratio loss with a moment matching mechanism for optimization and model selection. Experimental results show that GCHP can significantly reduce training time and the likelihood ratio loss with interarrival time probability assumptions can greatly improve the model performance.  \footnote{The source code is available at \href{https://github.com/ltz0120/Graph-Convolutional-Hawkes-Processes-GCHP}{https://github.com/ltz0120/Graph-Convolutional-Hawkes-Processes-GCHP}.}
	\end{abstract}

	
	
	
	\maketitle
	\section{Introduction}
	\textit{Attributed event sequences} are one of the most commonly encountered data objects in real-world applications. An attributed event sequence contains not only the timestamps of asynchronously generated events but also event features/attributes.\footnote{In this paper, we interchangeably use these two terms. Sometimes they are also referred to as ``marks'' in the literature of stochastic processes.} It is naturally generated from databases and event logfiles, and has been applied to various application scenarios and disciplines including financial transactions \cite{bacry2015hawkes}, natural language processing \cite{seonwoo2018hierarchical}, and spatial dependence among trees \cite{penttinen1992marked}, etc. Existing methods that deal with attributed event sequences are usually based on the statistical tool---marked point processes, which have been used for recommendation \cite{du2015time}, network inference \cite{linderman2014discovering},  fake news mitigation \cite{farajtabar2017fake}, and many other tasks \cite{yang2013mixture, porter2012self, tran2015netcodec}.
	
	To endow the probabilistic methods with better flexibility and effectiveness, some researchers \cite{du2016recurrent, xiao2017modeling, mei2017neural, omi2019fully} have explored the idea of incorporating marked point processes with neural networks, especially recurrent neural networks (RNNs), as they are applicable to the sequential nature.  The recurrent architecture of these models, however, makes it difficult to be accelerated by parallel mechanisms. As an alternative, the attention mechanism  has been applied to the learning of point processes in recent studies \cite{zuo2020transformer, zhang2020self}. In addition to the recurrent and attentive network architectures, a graph-based neural point process model \cite{shang2019geometric} has been applied to consider the geometry structure of Hawkes processes. Despite these architectures are believed to be more effective and have better representational power, it is still not clear whether more complex network architectures will do better for learning attributed event sequences. The other drawback of the current neural marked point process models is that the aforementioned models are often designated to particular forms of the conditional intensity function. For example, the RMTPP model \cite{du2016recurrent} utilizes an exponential form, whereas NHPP \cite{mei2017neural} utilizes a sigmoid function. Despite the computational convenience that these assumptions bring about, the representational capability of neural networks are also restricted. Moreover, existing approaches often involve the Monte Carlo integration \cite{binder2012monte} for predicting the next event, which is rather time-consuming instead.

	In this paper, we are trying to answer one of the most fundamental questions regarding neural point processes: how can we improve the model performance? Can we get a better model by making the network architecture more complicated? A short answer is NO. Neural point process models often exhibit what we call the \textit{performance saturation} phenomenon --- performance of the model increases and then stagnates at a certain point, as we increase the number of the network.
	So what causes the performance saturation of the neural marked point processes? The reason is that, the generalization error of neural point processes can be decomposed into network estimation error, model specification error (inductive bias) and some irreducible error caused by the randomness of the ground truth model.  As we utilize more parameters in the network, which is equivalent to increase the dimension of network function space, the network estimation error can be reduced, but not the model specification error. This tells us an important fact regarding neural point processes: defining a good probabilistic structure of the point process, sometimes is more important than chasing after fancy network architectures.
	
	Based on this observation, we try to improve the neural marked point processes in two ways: architectures and loss functions. We compare architectures among recurrent, attentive, graph-based ones and the combinations among them.  We propose a novel temporal-graph-based neural marked point processes, called graph convolutional Hawkes process (GCHP), which can achieve similar performance as the start-of-the-arts, but takes much less training time. The model falls into the category of nonlinear marked Hawkes process with multiplicative kernels. We also apply a convenient likelihood ratio loss based on moment matching approach, which can avoid the high time complexity that the Monte Carlo integration in the traditional methods brings about. Instead of using the conditional intensity function, we directly considers the conditional distributions of the interarrival times, and link up the loss functions with conditional intensity functions and conditional distributions of the interarrival times.

	
	The \textbf{contributions} of this paper are summarized as below.
	\begin{itemize}[topsep=0pt,itemsep=2pt,partopsep=2pt, parsep=2pt, leftmargin=0.2in]
		
		\item \textbf{We introduce the performance saturation phenomenon in neural point processes for the first time.} In this paper, we describe the \textit{performance saturation} phenomenon that  performance of neural network stop increasing as the neural network gets more complicated and has more parameters, which is different than the \textit{double descent} \cite{nakkiran2019deep} phenomenon as in classical neural networks. We provide an explanation based the generalization error decomposition.
		\item \textbf{We propose a simple graph-based network architecture for neural point process.} We propose a simple method, called GCHP, which is based on the temporal graph of the event history and can be easily incorporated by graph convolutional networks. This method helps not only significantly reduce the training time, but also improve the performance of existing methods.  
		\item  \textbf{We present an easy-to-use and effective loss function based on likelihood ratio test.}  We directly consider the distribution of interarrival times instead of imposing a specific assumption on the conditional intensity function. We propose to use a likelihood ratio type loss function which take into account both the model complexity and the likelihood of observations. We link up the equivalence among loss functions and intensity functions. Experimental results show that the our method can significantly improve the prediction accuracy.
		
	\end{itemize}
	
	\section{Related Work}
	Existing works on marked point processes can be classified into non-neural and neural-based ones. 
	
	\textbf{Non-neural marked point processes.} Models related to non-neural marked point processes are usually from the perspective of traditional statistical learning \cite{zhou2013learning, wang2016isotonic, xu2016learning, 8594954, li2019thinning}. These works carry out improvements in terms of incorporating statistical techniques such as regularization and non-parametric methods. \citeauthor{zhou2013learning} \cite{zhou2013learning} introduces nuclear and rank norm to the likelihood of multi-dimensional Hawkes processes, so that the sparse and low-rank pattern of the infectivity matrix can be recovered.
	\citeauthor{xu2016learning} \cite{xu2016learning} imposes a more general assumption of the decay kernels, which uses a series of basis functions such as exponential and Gaussian. \citeauthor{wang2016isotonic} \cite{wang2016isotonic} modulates the intensity function by an additional nonlinear link function, in order to capture the nonlinear effects. 
	Another major development of marked Hawkes processes is Bayesian Hawkes processes \cite{yang2013mixture, seonwoo2018hierarchical, li2020tweedie}. These models are usually fused with mixture models, especially in the context of natural language processing. Representative works are the Dirichlet-Hawkes Processes proposed in \cite{du2015dirichlet, xu2017dirichlet}, which take into account both textual contents and temporal information. Both models assume a Dirichlet prior distribution for the parameters, and therefore they are applicable to clustering tasks. 
	More recently,  some works \cite{seonwoo2018hierarchical, li2020tweedie} propose hierarchical Bayesian Hawkes processes  to deal with continuous features associated with events. One common drawback in Bayesian Hawkes processes is the poor scalability. The inference process for a Bayesian model is relatively time-consuming, and thus neural-based methods are getting more and more attractive.
	
	\textbf{Neural-based marked point processes.}  An active research line is to learn point processes with neural networks. The RMTPP \cite{du2016recurrent} model views the intensity function as a nonlinear function of the history, and uses a recurrent neural network to learn a representation of influences from the event history. Experimental results show that the model has better performance in both model fitting and prediction than traditional methods. \cite{mei2017neural} proposes a neural Hawkes process model named NHPP, which considers the interactions between events. The IRNN model \cite{xiao2017modeling} uses an intensity recurrent architecture that synergistically models time series and event sequence, making it able to capture both background and history effect. All of the above methods define respective intensity functions to be a specific parametric form. The fully neural point process model (FulNN) \cite{omi2019fully} relaxes the assumption of a parametric intensity function, and uses a fully connected neural network to output the cumulative hazard functions, which avoids defining a  specific form of the intensity function. However, the model fails to consider the features associated with each event. The geometric Hawkes process (GeoHP) model \cite{shang2019geometric} treats the parameter estimation of a vanilla Hawkes process as a matrix completion problem, and uses graph convolutional recurrent neural networks \cite{monti2017geometric} to solve it. Note that though GCN layers are used in GeoHP, they are used for learning the user/item embeddings. The main architecture of GeoHP is still RNN. Besides, the intensity function of the model is linear with even fewer parameters than the vanilla Hawkes process. More recently, some studies \cite{zhang2019self, zuo2020transformer} investigate the incorporation of attention mechanism for Hawkes processes.
	
\begin{figure}[t]
		\centering
		\includegraphics[width=0.41\textwidth]{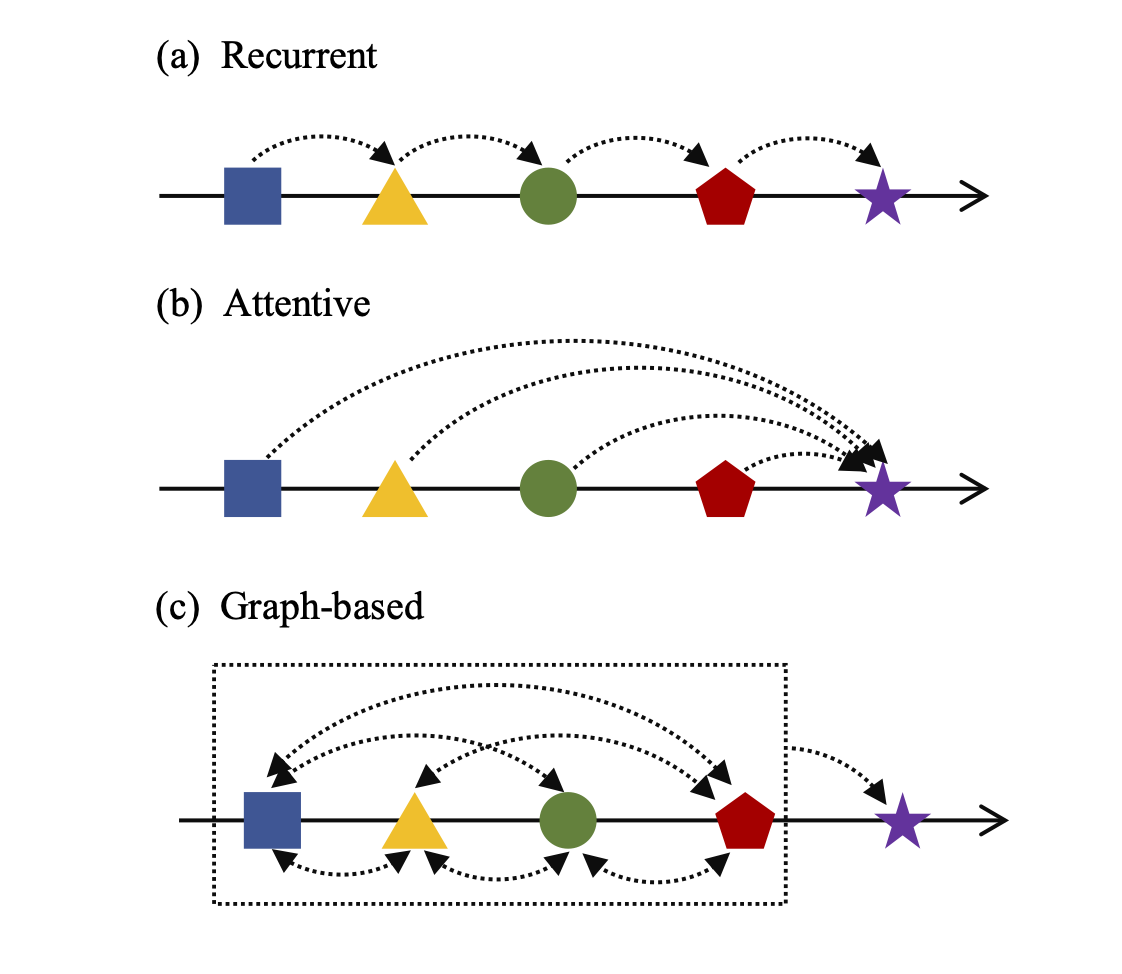}
		\caption{An illustration of the three message passing methods for neural point processes: (a) recurrent, (b) attentive, (c) graph-based network structure. The dashed lines with arrow heads denoted the direction of message passing. }
		\label{arch}
	\end{figure}

	Existing neural marked point processes can be categorized into three types in terms of the network architecture: recurrent, attentive and graph-based, as shown in Figure \ref{arch}. In Table \ref{table_comparison}, we summarize some important models in terms of the network architecture and the intensity function.

	\renewcommand{\arraystretch}{1.4}
	\begin{table}[tb]
		\centering
		\caption{Summary of some neural marked point processes.}\label{table_comparison}
		\resizebox{0.48\textwidth}{!}{
			\begin{tabular}{C{3.5cm} C{2.5cm} C{2.5cm}}
				\toprule[1.5pt]
				\textbf{Model} & \textbf{Network architecture}  & \textbf{Intensity function}   \\\midrule[1.5pt]
				
				RMTPP \cite{du2016recurrent}   & Recurrent  & Exponential \\\midrule[0.5pt]
				
				IRNN \cite{xiao2017modeling}  & Recurrent  & -- \\\midrule[0.5pt]
				Neural HP \cite{mei2017neural} & Recurrent  & Sigmoid \\\midrule[0.5pt]
				FulNN \cite{omi2019fully}    & Recurrent  & Softplus \\\midrule[0.5pt]
				GeoHP  \cite{shang2019geometric}  &  Graph-based, recurrent & Linear \\\midrule[0.5pt]
				Transformer HP \cite{zuo2020transformer} & Attentive & Softplus, exponential  \\  \midrule[0.5pt]
				Self-attentive HP \cite{zhang2020self} & Attentive  & Softplus \\
				\bottomrule[1.5pt]	
			\end{tabular}
		}
	\end{table}

	\section{Preliminaries}
	
	In this section, we briefly introduce two main preliminary techniques of our model. 

	\textbf{Marked point processes.} Marked point processes (MPP) \cite{daley2003introduction} are commonly used for modeling the temporal dynamics of attributed event sequences. A marked Hawkes process is a point process $\mathcal{N}(\cdot,\cdot)$ on $\mathcal{T} \times \mathcal{M}$, where $\mathcal{T} = [0, T]$ is the observation window and $\mathcal{M}$ the mark (feature) space. It is worth noting that if $\mathcal{M}$ is finite discrete, $\mathcal{N}$ is degenerated to a multi-dimensional Hawkes processes. In this paper, we assume that $\mathcal{M}$  can be continuous, i.e., $\mathcal{M}=\mathbb{R}^p$.  The continuous assumption is more general and common in real world.  \textit{Spatio-temporal Hawkes processes} \cite{reinhart2018review} are a good example of continuous mark space, as the location of a point (latitude and longitude) is in $\mathbb{R}^2$.
	Given the \textit{nature history} $\mathcal{H}_{t-}$, which is defined by the $\sigma$-algebra: $\mathcal{H}_{t-} = \sigma\{\mathcal{N}(s,\mathcal{M}; \omega):0<s<t\}$, where $\omega$ is a sampled path, the \textit{conditional intensity function} of a marked point process is defined by
	\begin{align*}
		\small
		\lambda(t, m \vert  \mathcal{H}_{t-}) = 
		\lim\limits_{\Delta_t, \Delta_m \to 0}\dfrac{ \mathbb{E} \big[  \mathcal{N} \left([t, t+\Delta_t) \times B(m, \Delta_m)\right) \vert \mathcal{H}_{t-}\big] }{\Delta_t \vert B(m, \Delta_m) \vert}, 
	\end{align*}
	where $\vert B(m, \Delta_m) \vert$ is the Lebesgue measure of the ball $B(m, \Delta_m)$ with radius $\Delta_m$. It can be decomposed by \cite{daley2003introduction}
	\begin{equation}
		\label{decomposition}
		\lambda(t, m \vert \mathcal{H}_{t-}) = \lambda_g(t\vert \mathcal{H}_{t-}) p(m \vert t, \mathcal{H}_{t-}),
	\end{equation}
	where $\lambda_g(t|\mathcal{H}_{t-})$ is the marginal intensity w.r.t. time, often referred to as the \textit{ground intensity}. The conditional marked and ground intensity function is often abbreviated to $\lambda^*(t, m)$ and $\lambda^*_g(t)$, respectively ,where the notation $*$  represents the intensity function is conditioned on the history $\mathcal{H}_{t-}$. 
	$p(m \vert t, \mathcal{H}_{t-})$ is the conditional mark density which refers to the distribution to be anticipated at the end of a time interval, not immediately after the next interval has begun. 
	Given a realization of attributed event sequence $\{(t_i, m_i):  i= 1,\dots, N\}$, the log-likelihood function is given by \cite{daley2003introduction}
	\begin{align*}
		\small
		\ell = \sum_{i=1}^{N}  \log \lambda_g(t_i\vert \mathcal{H}_{t_i-} ) - \int_{t_{i-1}}^{t_i} \lambda_g(t\vert \mathcal{H}_{t_i-}) dt + \log p(m_i\vert t_i,\mathcal{H}_{t_i-} ).
	\end{align*}
	
	\begin{figure*}[tb]
		\centering
			\vspace{-0.3in}
		\includegraphics[width=0.9\textwidth]{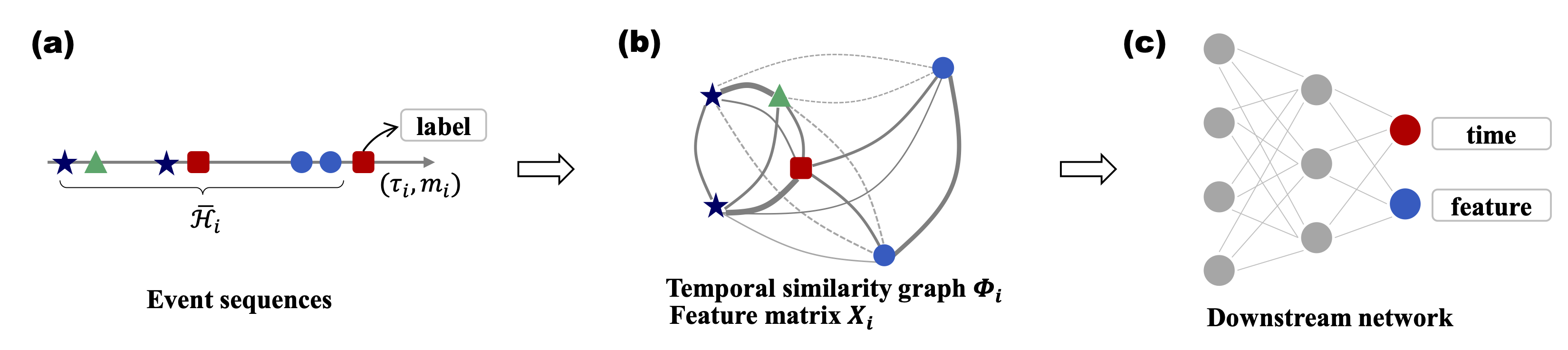}
		\caption{An illustration of the modeling flow of graph convolutional Hawkes processes (GCHP). (a)$\rightarrow$(b): transform into the attributed graph $(\Phi_i, X_i)$. (b)$\rightarrow$(c): input the data into the GCHP model. }
		\label{GCHP}
	\end{figure*}
	
	\textbf{Graph convolutional networks.} In recent years, GCNs  \cite{kipf2017semi, defferrard2016convolutional} have obtained great success as an efficient and effective model for graph-structured data. 
	Given an input graph with an adjacency matrix $\mathcal{A}$ and a feature matrix $\boldsymbol{X}$, GCNs encode both the topological information and the node attributes and produce an output with node embeddings.  
	The most representative model applies the new layer-wise propagation rule \cite{kipf2017semi}:
	\begin{equation*}
		\small
		H^{(l+1)} = \sigma(\tilde{A}H^{(l)}W^{(l)}),
	\end{equation*}
	where $\tilde{A}$ is a normalized adjacency matrix, $H^{(l)}$ is the output of the $l$-th layer,  $W^{(l)}$ is a layer-specific trainable weight matrix, and $\sigma$ is a non-linear activation function.
	

	\section{A simple temporal-graph-based architecture for neural marked point processes}

	In this section, we present a simple temporal-graph-based architecture for learning marked point processes. This method, which just utilizes graph convolutional layers, is easy and convenient to implement, and achieves as good performance as existing methods with much less training time. It can be viewed as a special case of a nonlinear marked Hawkes process with multiplicative kernel, therefore, we refer to our model as graph convolutional Hawkes processes (GCHP).
	
	\textbf{The model.} 
	Figure \ref{GCHP} illustrates the overarching modeling process of our GCHP method. We first scan the input attributed event sequence. For each event $(t_i, m_i)$, we obtain its trimmed history $\bar{ \mathcal{H}_{i}}$ with a preset number of prior events. We then transform the trimmed history $\bar{ \mathcal{H}_{i}}$ into a temporal similarity graph $\Phi_i$ and a feature matrix $\boldsymbol{X}_i$, which are then passed to graph convolutional layers.  Unlike \cite{du2016recurrent, mei2017neural} that assume a specific form of the intensity function, we use a moment matching strategy to approximate the intensity. To be specific, our GCHP model with two graph convolutional layers can be written as
	\begin{equation*}
		\left\{
		\begin{array}{l}
			\widehat{\tau}_i,\ \ \widehat{m}_i = F(H_i^{(2)}, \tilde{\Phi}_{i}),\\[2mm]
			H_i^{(2) } = \sigma(\tilde{\Phi}_{i}H_i^{(1)} W^{(1)} ),  \\[2mm]
			H_i^{(1) } = \sigma(\tilde{\Phi}_{i}\boldsymbol{X_{i}}W^{(0)} ), 
		\end{array}
		\right.
	\end{equation*} 
	where $\tilde{\Phi}_i= D_i^{-1/2}\Phi_i D_i^{-1/2}$, and $D_i$ is diagonal matrix of the degrees of $\Phi_i$. ``$:$'' denotes the concatenation of the two matrices. $F$ denotes fully connected layers, and $\sigma$ is an activation function, such as the ReLU.

	\textbf{The construction of temporal similarity graph  $\Phi$.} The temporal similarity graph $\Phi$ plays a crucial role in our model. As suggested by its name, it measures the similarity between events in the time domain. The use of temporal similarity graph is not arbitrary. It is essentially an important component in the intensity function of the nonlinear marked Hawkes processes with multiplicative kernels. 
	
	Given a symmetric kernel $\phi$, the weight between two events $(t_i, m_i)$ and $(t_j, m_j)$ can be defined by $\phi(t_i-  t_j)$.  
	The dimension of $\Phi$ is determined by the length of the trimmed history, which is preset by fixing the influential range, i.e.,  the number of past events that the next event is relevant to. Presetting the range makes the temporal similarity graphs and feature matrices of different events aligned. Trimmed history is also considered a reasonable approximation of the full history as the major influence comes from the closest events due to the decay of influence.
	
	\textbf{Nonliear marked Hawkes process with multiplicative kernels.} 
	We define a special type of marked Hawkes processes that incorporates multiplicative kernels for time and marks, whose intensity can be written as
	\begin{equation}
		\small
		\label{mhp}
		\lambda(t, m) = \mu p(m) + \int_{0}^{t} \int_{\mathcal{M}} \big( \phi \kappa \big) \ast dN,
	\end{equation}
	where $\mu$ is the base intensity and $p$ is a deterministic density function w.r.t. the mark $m$. $\phi $ and $\kappa $ are two positive definite kernel functions for arrival time and marks. $\ast$ denotes the convolution operation. It can be seen that such intensity is a linear convolution function. To relax the assumption of the linearity of intensity function, Eq. \eqref{mhp} can be extended to nonlinearity:
	\begin{equation}
		\small
		\label{NMHP}
		\lambda(t, m) = h \left( \mu p(m) + \int_{0}^{t} \int_{\mathcal{M}} \big( \phi \kappa \big) \ast dN \right),
	\end{equation}
	where $h: \mathbb{R} \rightarrow \mathbb{R}^+ $ is a non-negative  function. 
	It can be verified that the likelihood of such process can be viewed as a function of matrices $\Phi$ and $\mathcal{K}$. The former matrix is composed of $\phi(t_i-t_j)$'s. We call it the \textit{temporal similarity graph}, as it measures the similarity between each two events. 
	It can be seen that the feature kernel $\mathcal{K}$ provides an embedding method for the marks, in accordance with the theory of reproducing kernel Hilbert space. Therefore, the estimation of the next interarrival time $\tau$ and mark $m$, which is calculated from the estimated parameters by maximizing the likelihood, can be viewed as a function (denoted by $g$) of $\Phi$ and  $\mathcal{K}$.
	The estimation of the next interarrival time $\hat\tau$ and mark $\widehat{m}$ can be written by
	\begin{equation*}
		\hat\tau,\ \  \widehat{m}  = g (\Phi \odot \mathcal{K} ),
	\end{equation*}
	where $\Phi $ is the temporal similarity graph,  $\mathcal{K}$  the Gram matrix of the features (marks) and $\odot$ denotes the Hadamard product. The process can be interpreted in a sense that the next event is determined by the topology of the temporal similarity graph and the similarity of features. The closer two events are, the more similar their corresponding features will be.

	
	\renewcommand{\arraystretch}{1.0}
	\begin{table*}[tb]
		\centering
		\caption{Equivalence among conditional distributions, loss functions, and conditional intensity functions. $\Phi$ and $\Gamma$ are the cdf of a standard normal distribution and an upper incomplete gamma function, respectively. }\label{table_mm} 
		\resizebox{0.85\textwidth}{!}{
			\begin{tabular}{L{3.1cm} C{6.1cm} C{6.2cm}}
				\toprule[1.5pt]
				\textbf{Distribution} $p(\tau_i|\mathcal{H}_i)$  & \textbf{Equivalent loss function} $\ell_t (\tau_i, \hat{\tau}_i)$ & \textbf{Conditional intensity functions} $\lambda({\tau}_i|\mathcal{H}_i)$   \\ \midrule[0.5pt] 
				
			\multirow{2}{*}{\parbox{3.1cm}{	Exponential($\lambda$) }}& \multirow{2}{*}{\parbox{6.1cm}{ \centering $\sum_{i=1}^{N} \dfrac{\tau_i}{\hat{\tau}_i} + \log{\hat{\tau}_i}$ }}  & \multirow{2}{*}{\parbox{6.1cm}{ \centering $1/\hat{\tau}_i$ }} \\[0.15in] \midrule[0.5pt]
				
				\multirow{3}{*}{\parbox{3.1cm}{Gaussian($\mu$, $\sigma^2$)}} & 	\multirow{3}{*}{\parbox{6.1cm}{ \centering $\dfrac1\sigma_i\sum_{i=1}^{N} (\hat{\tau}_i-\tau_i )^2 -  2N\log \sigma_i $ }}&	\multirow{3}{*}{\parbox{6.2cm}{\centering $\dfrac{\exp{ \left( (\hat{\tau}_i-\tau_i )^2/2\sigma_i^2\right)}}{\sqrt{2\pi}\sigma\left(1-\Phi((\hat{\tau}_i-\tau_i)/\sigma_i\right)}$}}\\ 
				
				& &  \\ & &  \\\midrule[0.5pt]
				
				\multicolumn{1}{l}{\multirow{4}{*}{\parbox{3.1cm}{ Gamma($k$, $\theta$) }}} & \multirow{4}{*}{\parbox{6.1cm}{ \centering $\sum_{i=1}^{N} \log \Gamma(\dfrac{\hat\tau_i}{\theta_i}) + \dfrac{\tau_i}{\theta_i} - \dfrac{\hat\tau_i}{\theta_i} \log \dfrac{\tau_i}{\theta_i} $}} & \multirow{4}{*}{\parbox{6.2cm}{ \centering  $ \dfrac{\left(\frac{{\tau}_i}{\theta_i}\right)^{\left(\frac{{\hat\tau}_i}{\theta_i}\right)} \tau^{-1} e^{-\frac{{\tau}_i}{\theta_i}}}{\Gamma\left(\frac{\hat\tau_i}{\theta_i}, \frac{{\tau}_i}{\theta_i}\right)}$ }} \\
				& &  \\ & &  \\  & &  \\\midrule[0.5pt]
				
			\multirow{6}{*}{\parbox{3.1cm}{Laplacian($\mu$, $\sigma$) }}	 & 	\multirow{6}{*}{\parbox{6.1cm}{\centering $\dfrac1\sigma_i\sum_{i=1}^{N} \vert \hat{\tau}_i-\tau_i \vert$  }}  & \multirow{6}{*}{\parbox{6.1cm}{\centering $\begin{cases}
					\dfrac{1}{2\sigma_i\exp\left(-\dfrac{\hat{\tau}_i-\tau_i }{\sigma_i} \right) - \sigma_i}, \ \ \ \  \hat{\tau}_i \leq \tau_i \\
					\\
					\dfrac{1}{\sigma_i}\exp\left( -\dfrac{-2(\hat{\tau}_i-\tau_i) }{\sigma_i} \right), \ \ \ \  \hat{\tau}_i > \tau_i
				\end{cases}$ }} \\
				& &  \\ & &  \\  & &  \\ & &  \\ & &  \\	\bottomrule[1.5pt]	
			\end{tabular}
		}
	\end{table*}
	
	\textbf{Complexity analysis.}  Our model has a running time complexity of  $\mathcal{O}(Nm^2p)$ for each epoch, where $N$ is the number of events, $m$ and $p$ are the length of the trimmed history and the dimension of features, respectively. Note that $m \ll N$. For long sequences when $m \ll N$ does not hold, the temporal similarity graph (shown in Figure \ref{GCHP}) becomes sparse with the similarity values concentrate around diagonal entries. Therefore, the complexity of our model becomes $\mathcal{O}(Nmp)$ for long histories. This complexity is superior to \cite{zhou2013learning, zhou2013learningb, xu2016learning} whose complexity is $\mathcal{O}(N^3p)$. It is also better than \cite{achab2017uncovering}, which has a complexity of $\mathcal{O}(Np^2)$  for high-dimensional Hawkes processes where $p \gg m>0$. The THP \cite{zuo2020transformer} has the complexity of $\mathcal{O}(Nm^2p)$, which is much worse than our model for long histories. 
	

	\section{Likelihood Ratio and loss function}
	
	
	Generally, learning a stochastic process by maximizing the log-likelihood is viewed as an unsupervised task. For neural point processes, 
	each event $(t_i, m_i)$ in the input sequence is treated as a label.  The general objective function can be written as
	\begin{equation*}
		\small
		\text{loss} =  \sum\limits_{i=1}^{N}\left( \ell_m(m_i, \widehat{m}_i ) + c \ell_t (\tau_i, \hat{\tau}_i)\right).
	\end{equation*}
	Here $\widehat{m}_i $ and $\hat{\tau}_i $ are the outputs for the feature and interarrival time of the $i$-th event, given the history $\bar{ \mathcal{H}_{i}}$. $\tau_i = t_i - t_{i-1}$, $t_0=0$ and $m_i$ is the actual interarrival time and feature of the next event. 
	$c$ is a hyper-parameter controlling the weight of time. $\ell_m$ and $\ell_t$ are the respective loss functions. 
	It is worth noting that, the maximum likelihood estimator of neural point processes also admits the form, as a result of the invariance property. 
	
	One of the most challenging parts in applying marked point processes is that the exact form of the conditional intensity $\lambda^*(t, m)$ is not known. Traditionally, the loss function is designed by assuming a specific form of the conditional intensity function. There are several attempts made by researchers to designate some specific forms for the intensity. RMTPP \cite{du2016recurrent} uses an exponential form, whereas NHP \cite{mei2017neural} adopts sigmoid. Such choices, however, may restrict the expressive power of neural networks. Moreover, the calculation of the expectation of the next interarrival time usually does not have analytic solutions, and thus one has to turn to numerical methods, such as Monte Carlo simulation, which is computationally unfriendly. Recently, \cite{omi2019fully} proposes an approach that avoids the specification of the intensity. It first models the integral of the intensity using a feedforward neural network and then obtains the intensity function as its derivative. However, this method is unable to perform long-term predictions, as the derivatives for future events are not available.  In this paper, we propose to consider the distribution of the interarrival times, instead of the intensity function, as a result the Monte Carlo integration can be avoided. The next lemma states that the probabilistic structure of a point process can be equivalently defined by  its conditional intensity function as well as the conditional density of the interarrival times. 
	
	\begin{lemma}[Equivalence between conditional intensity function and conditional density of the interarrival times \cite{daley2003introduction}]
		A regular point process is specified uniquely by the conditional intensity function $\lambda_i^*(t)$ if and only if the conditional probability densities of the next arrival time satisfy that $p_i^*(t) = \lambda_i^*(t) \exp \left\{-\int_{t_{i-1}}^{t} \lambda_i^*(s)ds\right\}$, for all $i=1, 2, \cdots.$
	\end{lemma}	
	The proof of this lemma can be found in \cite{daley2003introduction}. This result tells that we do not have to designate the form of the conditional intensity function, but instead we can directly impose the probabilistic assumptions on the interarrival times.
	
	\textbf{Likehihood ratio loss function}. As we would like to jointly optimize the network structure and the parameters, we propose a loss function based on the likelihood ratio statistic, which takes into account both the likelihood of the observations and the complexity (in terms of the number of free parameters in a network) of the model. The optimal network structure and parameter can be obtained by optimizing the likelihood ratio loss function, which can by written as,
	\begin{align*}
		\boldsymbol w^*, \boldsymbol \Xi^* = \arg \inf\limits_{\boldsymbol w, \boldsymbol \Xi \in \mathbb{F} } \inf\limits_{p \in \mathbb{P}} \log \dfrac{\sup\limits_{\boldsymbol{\theta} \in \boldsymbol{\Theta}_0} \prod_{i=1}^{N}p(\tau_i, m_i | \boldsymbol{\theta})}{\sup\limits_{\boldsymbol{\theta} \in \boldsymbol{\Theta}_1} \prod_{i=1}^{N} p_i(\tau_i, m_i |\mathcal{H}_i; \boldsymbol{\theta}) \chi^2_\alpha(N-d_{\boldsymbol \Xi})}
	\end{align*}
	where
	$\boldsymbol{\Theta}_0$ is the null parameter space, $\boldsymbol{\Theta}_1$ the parameter space restricted by the neural network, $\mathbb{P}$ the distribution family, $\boldsymbol{w}$ the network parameter, and $\boldsymbol \Xi$ the network structure. According to the traditional model selection theory, this likelihood ratio has a $\chi^2$ distribution with degree of $N-d_\Xi$, where $N$ is the number of samples, and $d_\Xi$ is the number of free parameters in model $d_\Xi$. Therefore, we introduce a $\chi^2$ coefficient in the loss function, where $\chi^2_\alpha(N-d_{\boldsymbol \Xi})$ represents the $\alpha$-percentile of the $\chi^2$ distribution with degree $N-d_\Xi$. This loss function measures the goodness-of-fit, and it can be reduced to a generalized likelihood ratio test problem.

	\textbf{Moment matching.}  We propose to use a moment matching mechanism in the loss function. We directly output the expectation of the next interarrival time $\hat\tau_i = \mathbb{E}(\tau_i \vert \bar{ \mathcal{H}_{i}} )$ and feature $\hat m_i = \mathbb{E}(m_i \vert \bar{ \mathcal{H}_{i}} )$, which are the first-order moment of the next interarrival time and mark, respectively. This method has two benefits. First, it is convenient for the network to predict the next event. Second, it reduces the number of free parameters to estimate, making the model more robust to overfitting.

	\textbf{Discussion.} It is worth noting that this model selection method can be regarded as a traditional statistical goodness-of-fit problem. It is only valid for the classical underparameterized situations where the number of the free parameters in the network is smaller than the number of events for training.
	
	
	\begin{figure}[tb]
		\centering
			\vspace{-0.3in}
		\includegraphics[width=0.5\textwidth]{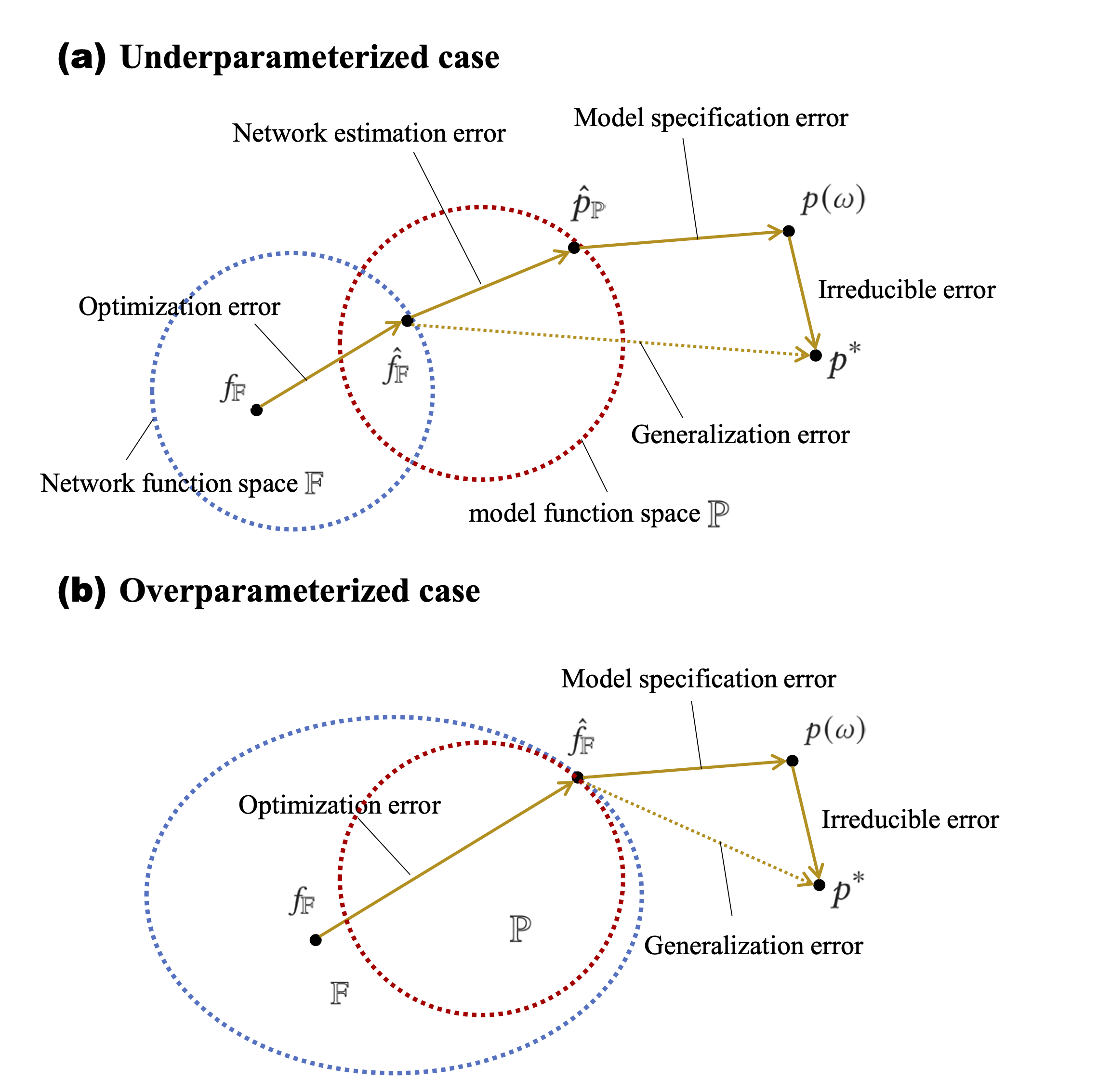}
		\caption{An illustration of the generalization error decomposition and the performance saturation phenomenon. 
			$\mathbb{F}$ and $\mathbb{P}$ are the function spaces defined by the neural network and the point process probabilistic assumptions.  $f_{\scaleto{\mathbb{F}}{4pt}}$ is a network model with arbitrary parameter. $\hat f_{\scaleto{\mathbb{F}}{4pt}}$ represents the optimal estimation learned by neural network, and $\hat p_{\scaleto{\mathbb{P}}{4pt}}$ denotes the optimal estimation of under the probabilistic assumption. $p^*$ and $p(\omega)$ indicate the ground truth and a realization, respectively. (a) In the underparameterized case, network function space $\mathbb{F}$ is not large enough to find the optimal estimation given the probabilistic model restrictions. (b) In the underparameterized case, $\mathbb{F}$ is large enough to obtain the optimal estimation, where the network estimation error is eliminated. However, the performance cannot be further improved as the model specification error cannot be reduced by modifying the neural network only.  }
			\vspace{-0.1in}
		\label{error}
	\end{figure}
	
	\section{The saturation phenomenon for the learning of neural point processes}

	The neural networks for a probabilistic model can be viewed as an  estimator for the unknown parameters. Classical learning theory \cite{hastie2009elements} indicates as the number of parameters in a model increases, the model becomes prone to overfitting and the test error gets larger.  Recent studies \cite{ nakkiran2019deep}  find that many deep learning tasks exhibit a ``double descent'' phenomenon where model performance initially gets worse and then gets better as the number of free parameters in the model increases. Therefore, many believes that a mammoth neural network model always means a good opportunity to obtain  better performance. However, this conjecture is not valid when it comes to neural marked point processes. Instead, the performance of neural networks often becomes ``saturated'' -- no matter how much efforts are put into making the neural network more complicated and has more parameters, the performance just stop increasing. 
	
	The reason of this phenomenon is that the representation capability of the neural network is capped by the assumptions of point process. An extreme example would be a neural homogeneous Poisson process, which is like ``to break a butterfly upon a wheel'' -- no matter how delicate the network is, its generalization ability is still rather weak. We present an explanation in Figure \ref{error}. It can be seen that the generalization error can be decomposed into three parts: network estimation error, model specification error and some irreducible error caused by the randomness of the ground truth. In the underparameterized case, the function space defined by the neural network, ie. $\mathbb{F}$,  may not include the function space defined by the point process probability structure, ie. $\mathbb{M}$, leading to the network estimation error. This error can be reduced by making the network to be more complicated and have more parameters, until the optimal estimation $\hat p_{\scaleto{\mathbb{P}}{4pt}}$ is included in $\mathbb{F}$. After that, however, all the efforts put into enlarging $\mathbb{F}$ are all in vain, as $\hat p_{\scaleto{\mathbb{P}}{4pt}}$ is already achieved. As a result, the ``double descent'' phenomenon does not occur when it comes to neural point processes.
	
\begin{figure}[tb]
		\centering
		\vspace{-0.1in}
		\includegraphics[width=0.48\textwidth]{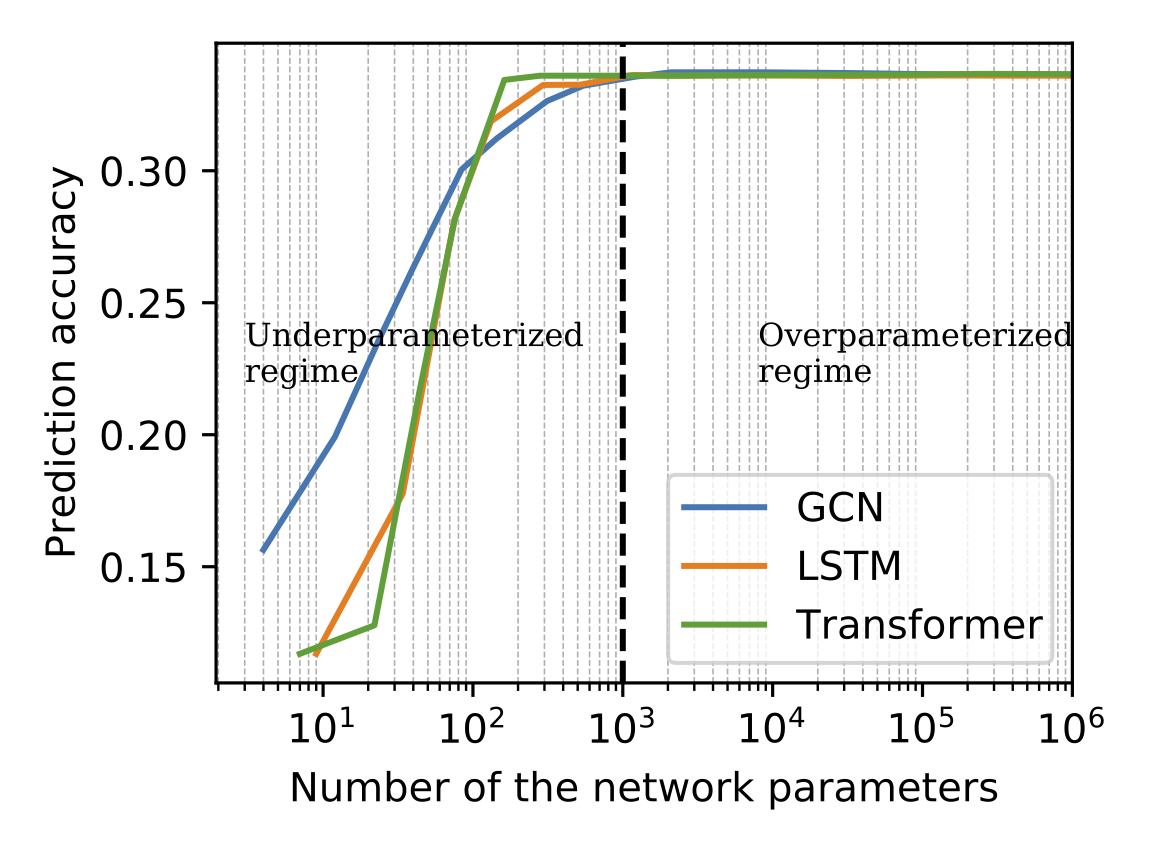}
		\caption{The performance saturation phenomenon for neural point processes.  It can be seen that the model performance on test dataset stagnates after as the number of network parameters increases.}
		\label{dim}
	\end{figure}

	We perform an example experiment on a synthetic dataset. The dataset is simulated from a 10-dimensional Hawkes process. The description of the dataset can be found in Section 7. We present the accuracy of mark prediction on test dataset using three different network structures in Figure \ref{dim}. When the network is underparameterized, the performance continuously increases as the size of network expands, until the performance reaches certain point. In the overparameterized regime, all the networks have similar performance. Neither descent nor ``double ascent'' is observed. 
	
\section{Experiments}
In this section, we evaluate our model against some state-of-the-art baselines on one synthetic and threes real-world datasets.

\textbf{Datasets.} The datasets  we use are listed as follows. We summarize the statistics of the datasets in Table \ref{table_statistics}.
\begin{itemize}
	\item[$-$] \textit{Hawkes}: a synthetic dataset with categorical features. 
	The event sequences are generated from a 10-dimensional Hawkes process with uniformly sampled parameters. 
	\item[$-$] \textit{IPTV} \cite{luo2015multi}: a real-world dataset with categorical features. The dataset consists of IPTV viewing events with timestamps and categories. 
	\item[$-$] \textit{Weeplace} \cite{liu2013learning}: a real-world dataset with both categorical and continuous features. The dataset contains the check-in histories of users at different locations (longitudes and latitudes). 
	\item[$-$] \textit{ATM} \cite{xiao2017modeling}: a real-world dataset with categorical features. The dataset is composed of the event logs of error reporting and failure tickets.
\end{itemize}

\renewcommand{\arraystretch}{1.2}
\begin{table}[tb]
	\centering
	\caption{Statistics of datasets.}\label{table_statistics}
	\vspace{-0.1in}
			\resizebox{0.48\textwidth}{!}{	
				\begin{tabular}{p{1cm} c c c c c}
					\toprule[1.5pt]
					\multicolumn{1}{c}{\multirow{2}{*}{\textbf{Dataset}}} & \multicolumn{2}{c}{\multirow{1}{*}{\textbf{\# of events}}} & \multicolumn{2}{c}{\multirow{1}{*}{\textbf{\# of sequences}} }& \multirow{2}{1.5cm}{\textbf{\# of event types $K$}} \\ \cline{2-5}
					& Train set &  Test set & Train set & Test set & \\ \midrule[1pt]
					Hawkes & 36k  & 7k & 100 & 40 & 10 \\
					ATM & 370k & 182k & 1085 & 469 & 7  \\ 
					Weeplace & 98k & 31k & 21 & 8 & 8  \\ 
					IPTV &731k &243k &227& 75& 16 \\
					\bottomrule[1.5pt]	
			\end{tabular}}
\end{table}

\textbf{Experimental environment.}  All the experiments were conducted on a server with 64G RAM, a 16 logical cores CPU (AMD Ryzen Threadripper 1900X) and 4 GPUs (Nvidia GeForce GTX 1080 Ti) for acceleration.

\begin{table}[tb]
	\centering
	\vspace{-0.1in}
	\caption{Performance on prediction.}\label{table_large}
	\renewcommand{\arraystretch}{1.05}
	\resizebox{0.5\textwidth}{!}{	
		\begin{tabular}{c p{1.5cm} p{1.2cm} p{1.2cm} c}
			\toprule[2pt]
			
			\multicolumn{1}{c}{\multirow{3}{*}{\textbf{Dataset}}} &	\multicolumn{1}{c}{\multirow{3}{*}{\textbf{Model}}} & \multicolumn{1}{c}{\multirow{3}{2cm}{\centering \textbf{Accuracy \\ (feature)}}} & \multicolumn{1}{c}{\multirow{3}{1.8cm}{\centering \textbf{RMSE \\ (time)}}}& \multirow{3}{1.5cm}{\textbf{Average \\ 
					Running time (s)}}  \\ 
			& & & \\
			& & & \\
			\midrule[1pt]

			\multicolumn{1}{l}{\multirow{10}{*}{\parbox{2cm}{\centering Hawkes } }} 
			&	\multicolumn{1}{l}{\multirow{1}{*}{\parbox{2cm}{\centering RMTPP \cite{du2016recurrent}} }} & \multicolumn{1}{c}{\multirow{1}{*}{ 32.46\%}  }   &    \multicolumn{1}{c}{ 5.565} & 0.451
			\\
			&	\multicolumn{1}{l}{\multirow{1}{*}{\parbox{2cm}{\centering IRNN \cite{xiao2017modeling}} }}  &\multicolumn{1}{c}{\multirow{1}{*}{ 33.40\%}  }  &   \multicolumn{1}{c}{ 4.395} &  0.475\\
			&  	\multicolumn{1}{l}{\multirow{1}{*}{\parbox{2cm}{\centering NHPP \cite{mei2017neural}}}} &  \multicolumn{1}{c}{\multirow{1}{*}{ 33.61\%}} & \multicolumn{1}{c}{ 4.480 } & 46.47 \\
			&  	\multicolumn{1}{l}{\multirow{1}{*}{\parbox{2cm}{\centering MAHP \cite{xu2018superposition}}}} & \multicolumn{1}{c}{\multirow{1}{*}{ 10.01\%}} & \multicolumn{1}{c}{  4.898 } & 1.794 \\
			&  	\multicolumn{1}{l}{\multirow{1}{*}{\parbox{2cm}{\centering GeoHP \cite{shang2019geometric}}}} & \multicolumn{1}{c}{\multirow{1}{*}{ 22.91\%}} & \multicolumn{1}{c}{  12.62 } &  38.94 \\
			&
			\multicolumn{1}{l}{\multirow{1}{*}{\parbox{2cm}{\centering THP \cite{zuo2020transformer}}}}   & \multicolumn{1}{c}{\multirow{1}{*}{ 33.27\%}} & \multicolumn{1}{c}{ 35.01  } & 122.7  \\
			\cmidrule{2-5}
			&
			\multicolumn{1}{l}{\multirow{1}{*}{\parbox{2cm}{\centering 1-layer GCN }}}   & \multicolumn{1}{c}{\multirow{1}{*}{ 33.75\%}} & \multicolumn{1}{c}{ 4.506  } & \underline{\textbf{0.0792}}  \\
			
			&	\multicolumn{1}{l}{\multirow{1}{*}{\parbox{2cm}{\centering 2-layer GCN}}} & \multicolumn{1}{c}{ \multirow{1}{*}{  33.81\%}}  &    \multicolumn{1}{c}{ \underline{\textbf{4.385}} } & 0.0888 \\
			&
			\multicolumn{1}{l}{\multirow{1}{*}{\parbox{2cm}{\centering GCN + LSTM }}}   & \multicolumn{1}{c}{\multirow{1}{*}{ 33.16\%}} & \multicolumn{1}{c}{ 4.374 } & 0.1258  \\
			&
			\multicolumn{1}{l}{\multirow{1}{*}{\parbox{2cm}{\centering GCN + TFM }}}   & \multicolumn{1}{c}{\multirow{1}{*}{ \textbf{\underline{ 33.97\%}}}} & \multicolumn{1}{c}{ 4.392  } & 0.2551  \\
			
			
			\midrule[1pt]
			
			\multicolumn{1}{l}{\multirow{10}{*}{\parbox{2cm}{\centering ATM }}} 
			&	\multicolumn{1}{l}{\multirow{1}{*}{\parbox{2cm}{\centering RMTPP \cite{du2016recurrent}}}} & \multicolumn{1}{c}{\multirow{1}{*}{ 76.64\%}  }   &    \multicolumn{1}{c}{7.150} &5.756  \\
			&	\multicolumn{1}{l}{\multirow{1}{*}{\parbox{2cm}{\centering IRNN \cite{xiao2017modeling}}}}  &\multicolumn{1}{c}{\multirow{1}{*}{ 76.19\%}  }  & \multicolumn{1}{c}{2.793} & 6.299 \\
			&  	\multicolumn{1}{l}{\multirow{1}{*}{\parbox{2cm}{\centering NHPP \cite{mei2017neural}}}} &  \multicolumn{1}{c}{\multirow{1}{*}{ 33.78\%}} & \multicolumn{1}{c}{ 7.558  } &  660.52  \\
			& \multicolumn{1}{l}{\multirow{1}{*}{\parbox{2cm}{\centering MAHP \cite{xu2018superposition}} }} & \multicolumn{1}{c}{\multirow{1}{*}{ 41.91\%}} & \multicolumn{1}{c}{ 3.202  } &  24.876  \\
			&  	\multicolumn{1}{l}{\multirow{1}{*}{\parbox{2cm}{\centering GeoHP \cite{shang2019geometric}}}} & \multicolumn{1}{c}{\multirow{1}{*}{ 14.91\%}} & \multicolumn{1}{c}{  9.268  } &  872.40 \\
			&  	\multicolumn{1}{l}{\multirow{1}{*}{\parbox{2cm}{\centering THP \cite{zuo2020transformer}}}} & \multicolumn{1}{c}{\multirow{1}{*}{ 68.76\%}} & \multicolumn{1}{c}{ 4.534 } &  14.612 \\
			
			\cmidrule{2-5}
			&
			\multicolumn{1}{l}{\multirow{1}{*}{\parbox{2cm}{\centering 1-layer GCN }}}   & \multicolumn{1}{c}{\multirow{1}{*}{ 76.56\%}} & \multicolumn{1}{c}{ 2.825  } & \underline{\textbf{0.2611}}  \\
			&	\multicolumn{1}{l}{\multirow{1}{*}{\parbox{2cm}{\centering 2-layer GCN }}} & \multicolumn{1}{c}{ \multirow{1}{*}{ 90.88\%}}  &    \multicolumn{1}{c}{\underline{ \textbf{2.612}}} & 0.3993  \\
			&
			\multicolumn{1}{l}{\multirow{1}{*}{\parbox{2cm}{\centering GCN + LSTM }}}   & \multicolumn{1}{c}{\multirow{1}{*}{ \underline{\textbf{91.41\%}}}} & \multicolumn{1}{c}{ 2.899 } & 0.5061  \\
			&
			\multicolumn{1}{l}{\multirow{1}{*}{\parbox{2cm}{\centering GCN + TFM }}}   & \multicolumn{1}{c}{\multirow{1}{*}{ 91.08\%}} & \multicolumn{1}{c}{ 2.767  } & 1.5754 \\
			
			\midrule[1pt]
			
			\multicolumn{1}{l}{\multirow{10}{*}{\parbox{2cm}{\centering IPTV } }} 
			&	\multicolumn{1}{l}{\multirow{1}{*}{\parbox{2cm}{\centering RMTPP \cite{du2016recurrent}} }} & \multicolumn{1}{c}{\multirow{1}{*}{  57.57\%}  }   &    \multicolumn{1}{c}{34.382}  & 11.281  \\
			&	\multicolumn{1}{l}{\multirow{1}{*}{\parbox{2cm}{\centering IRNN \cite{xiao2017modeling}} }}  &\multicolumn{1}{c}{\multirow{1}{*}{  58.63\%}  }   &    \multicolumn{1}{c}{34.311}   &  11.065\\
			&  	\multicolumn{1}{l}{\multirow{1}{*}{\parbox{2cm}{\centering NHPP \cite{mei2017neural}} }} & \multicolumn{1}{c}{\multirow{1}{*}{ 31.05\%}} & \multicolumn{1}{c}{  19.929 } &  1070.15 \\ 
			&  	\multicolumn{1}{l}{\multirow{1}{*}{\parbox{2cm}{\centering MAHP \cite{xu2018superposition}} }} &  \multicolumn{1}{c}{\multirow{1}{*}{ 18.02\%}} & \multicolumn{1}{c}{  36.738  } & 28.213\\
			&  	\multicolumn{1}{l}{\multirow{1}{*}{\parbox{2cm}{\centering GeoHP \cite{shang2019geometric}}}} & \multicolumn{1}{c}{\multirow{1}{*}{ 43.12\%}} & \multicolumn{1}{c}{ 25.421  } &  907.91 \\
			&  	\multicolumn{1}{l}{\multirow{1}{*}{\parbox{2cm}{\centering THP \cite{zuo2020transformer}}}} & \multicolumn{1}{c}{\multirow{1}{*}{ 71.94\%}} & \multicolumn{1}{c}{ 31.325  } &  10.031 \\

			\cmidrule{2-5}
			&
			\multicolumn{1}{l}{\multirow{1}{*}{\parbox{2cm}{\centering 1-layer GCN }}}   & \multicolumn{1}{c}{\multirow{1}{*}{ 75.28\%}} & \multicolumn{1}{c}{ 11.162  } & \underline{\textbf{1.8634}} \\
			
			&	\multicolumn{1}{l}{\multirow{1}{*}{\parbox{2cm}{\centering 2-layer GCN } }} & \multicolumn{1}{c}{ \multirow{1}{*}{ 75.35\%}}  &    \multicolumn{1}{c}{\underline{\textbf{10.866}}} & 2.0753 \\ 
			&
			\multicolumn{1}{l}{\multirow{1}{*}{\parbox{2cm}{\centering GCN + LSTM }}}   & \multicolumn{1}{c}{\multirow{1}{*}{ \underline{\textbf{76.11\%}}}} & \multicolumn{1}{c}{ 11.133 } & 3.1946 \\
			&
			\multicolumn{1}{l}{\multirow{1}{*}{\parbox{2cm}{\centering GCN + TFM }}}   & \multicolumn{1}{c}{\multirow{1}{*}{ 76.01\%}} & \multicolumn{1}{c}{ 11.139  } & 7.6419  \\
			
			\midrule[1pt]
			
			\multicolumn{1}{l}{\multirow{10}{*}{\parbox{2cm}{\centering Weeplace } }} 
			&	\multicolumn{1}{l}{\multirow{1}{*}{\parbox{2cm}{\centering RMTPP \cite{du2016recurrent}} }} & \multicolumn{1}{c}{\multirow{1}{*}{  22.07\%}  }   &    \multicolumn{1}{c}{7.162} &1.400  \\
			&	\multicolumn{1}{l}{\multirow{1}{*}{\parbox{2cm}{\centering IRNN \cite{xiao2017modeling}} }}  & \multicolumn{1}{c}{\multirow{1}{*}{  23.37\%}  }   &  \multicolumn{1}{c}{\underline{\textbf{6.448}}} &1.434 \\
			&  	\multicolumn{1}{l}{\multirow{1}{*}{\parbox{2cm}{\centering NHPP \cite{mei2017neural}} }} & \multicolumn{1}{c}{\multirow{1}{*}{ 25.71\%}} & \multicolumn{1}{c}{ 6.773  } & 140.26 \\
			&  	\multicolumn{1}{l}{\multirow{1}{*}{\parbox{2cm}{\centering MAHP \cite{xu2018superposition} } }} & \multicolumn{1}{c}{\multirow{1}{*}{ 15.13\%}} & \multicolumn{1}{c}{ 6.969  } &  5.210 \\
			&  	\multicolumn{1}{l}{\multirow{1}{*}{\parbox{2cm}{\centering GeoHP \cite{shang2019geometric}}}} & \multicolumn{1}{c}{\multirow{1}{*}{  17.74\%}} & \multicolumn{1}{c}{  28.28 } & 42.89 \\
			&
			\multicolumn{1}{l}{\multirow{1}{*}{\parbox{2cm}{\centering THP \cite{zuo2020transformer}}}}   & \multicolumn{1}{c}{\multirow{1}{*}{ 29.24\%}} & \multicolumn{1}{c}{ 51.78  } & 51.15 \\
			
			\cmidrule{2-5}
			&
			\multicolumn{1}{l}{\multirow{1}{*}{\parbox{2cm}{\centering 1-layer GCN }}}   & \multicolumn{1}{c}{\multirow{1}{*}{ 31.61\%}} & \multicolumn{1}{c}{ 6.498  } & \underline{\textbf{0.1831}}  \\
			&	\multicolumn{1}{l}{\multirow{1}{*}{\parbox{2cm}{\centering 2-layer GCN} }} & \multicolumn{1}{c}{ \multirow{1}{*}{   31.81\%}}  &    \multicolumn{1}{c}{6.493} & 0.2090  \\
			&
			\multicolumn{1}{l}{\multirow{1}{*}{\parbox{2cm}{\centering GCN + LSTM }}}   & \multicolumn{1}{c}{\multirow{1}{*}{ \underline{\textbf{32.09\%}}}} & \multicolumn{1}{c}{ 6.525 } & 0.2832  \\
			&
			\multicolumn{1}{l}{\multirow{1}{*}{\parbox{2cm}{\centering GCN + TFM }}}   & \multicolumn{1}{c}{\multirow{1}{*}{ 30.05\%}} & \multicolumn{1}{c}{ 6.563 } & 0.6990  \\

			\bottomrule[2pt]
	\end{tabular}}
		\vspace{-0.1in}
\end{table}

\subsection{Task 1: Comparison Among Network Structures}
In this task, we compare the performance of different neural network architectures, to demonstrate the advantage of applying GCN networks in marked point processes.

\textbf{Baselines.} We compare our model with six state-of-the-art neural-based methods: RMTPP \cite{du2016recurrent} (RNN-based model), IRNN \cite{xiao2017modeling}, NHPP \cite{mei2017neural} (LSTM-based model),  MAHP \cite{xu2018superposition} and GeoHP \cite{shang2019geometric}, and THP \cite{zuo2020transformer} (transformer-based model). Meanwhile, we also compare with some variants of our model including the one-layer GCN and the combinations of the GCN with LSTM and the GCN with transformer. 


\textbf{Metrics.} We assess the performance of each model in three aspects: time prediction, feature prediction and training time. We use \textbf{RMSE} for the time prediction, and we measure the categorical features prediction by the percentage of correct predictions (\textbf{Accuracy}). A higher accuracy and a lower RMSE indicate a better performance. 
The training time per epoch is also recorded, as a measure of the model's efficiency.

\textbf{Experimental settings.}
We apply likelihood ratio loss to train our model. The hyper-parameters of all models were tuned for the best performance. We use a single fully connected layer after the graph convolutional layers in our model, to predict the time and event category.

\textbf{Discussion.}  The experimental results are shown in Table \ref{table_large}. From the experimental results, we can observe that our GCN-based model outperforms the baseline methods in terms of time prediction error, category prediction accuracy and training time. In addition, by combining with GCN networks, the performance of LSTM and transformer are greatly improved. We contribute the performance improvements into three aspects: (1). The GCN model encodes the event correlations into the temporal similarity graph, which can better encode the relations among each event than other models. (2). The lightweight of GCN can greatly speed up the training processes. (3). The likelihood ratio loss fits the task much better than other intensity losses used in the baseline methods. We furthermore show this point in Task 2.


\subsection{Task 2: Comparison Among Loss Functions}

To test the model's performance under different optimization objective, i.e. loss function, we conduct experiments on the state-of-the-art models such as LSTM and transformer with different loss types. The experiment results are shown in Table \ref{loss_type}.


\textbf{Experimental settings.} We select some state-of-the-art models as the representatives of the combination of certain network architecture and loss type: RMTPP \cite{du2016recurrent} represents the RNN-based model with exponential intensity loss, NHPP \cite{mei2017neural} represents the LSTM-based model with softmax intensity loss, THP \cite{zuo2020transformer} represents the transformer-based model with exponential intensity loss. We further implement two models: the LSTM-based model which has one layer LSTM, and transformer-based model which has one layer transformer, based on our proposed likelihood ratio loss which minimizes the exponential interarrival loss(discussed in Section 5). We carefully tune the hyper-parameters to enable all the models achieve their best performance. 

\textbf{Discussion.} From Table \ref{loss_type}, we can infer that the optimization objective can greatly affect the performance of the model. With adopting our proposed likelihood ratio loss function, the prediction accuracy is consistently enhanced among four datasets, which demonstrates the significance of the assumption on the exponential distribution of interarrival times. In addition, the baselines \cite{du2016recurrent,mei2017neural,zuo2020transformer} apply Monte Carlo integration to approximate their intensity, which may slowdown the entire inference process. By adopting our exponential distribution assumption with moment matching, we can significantly speed up the training process.

\begin{table}[tb]
	\centering
	\caption{Performance comparison with different loss types. Exp interarrival represents the exponential distribution assumption on the interarrival times.  Exp/Sigmoid/Softplus intensity represents the respective form assumptions on the conditional intensity function.  }\label{table_pre_short}
	\renewcommand{\arraystretch}{1.1}
	\resizebox{0.45\textwidth}{!}{	
		\begin{tabular}{c p{1.2cm} p{1.2cm} c}
			\toprule[2pt]
			
			\multicolumn{1}{c}{\multirow{2}{*}{\textbf{Dataset}}} &	\multicolumn{1}{c}{\multirow{2}{3.5cm}{\centering \textbf{Model}}} & \multicolumn{1}{c}{\multirow{2}{1cm}{\centering \textbf{Accuracy \\ (feature)}}} & \multicolumn{1}{c}{\multirow{2}{1cm}{\centering \textbf{RMSE \\ (time)}}} \\ 
			& & \\
			\midrule[1pt]

			\multicolumn{1}{l}{\multirow{5}{*}{\parbox{1.5cm}{\centering Hawkes } }} &	\multicolumn{1}{l}{\multirow{1}{*}{\parbox{4cm}{\centering \textbf{LSTM + Exp interarrival}}}} & \multicolumn{1}{c}{ \multirow{1}{*}{  \textbf{\underline{ 33.68\%}}}}  &    \multicolumn{1}{c}{ \underline{\textbf{4.436}} } \\ 
			&	\multicolumn{1}{l}{\multirow{1}{*}{\parbox{4cm}{\centering RNN + Exp intensity \cite{du2016recurrent}} }} & \multicolumn{1}{c}{\multirow{1}{*}{ 32.46\%}  }   &    \multicolumn{1}{c}{ 5.565}
			\\
			&	\multicolumn{1}{l}{\multirow{1}{*}{\parbox{4cm}{\centering LSTM + Sigmoid  intensity \cite{mei2017neural}} }}  &\multicolumn{1}{c}{\multirow{1}{*}{ 33.61\%}  }  &   \multicolumn{1}{c}{ 4.480}\\\cmidrule{2-4}
			&  	\multicolumn{1}{l}{\multirow{1}{*}{\parbox{4cm}{\centering \textbf{TFM + Exp interarrival} }}} &  \multicolumn{1}{c}{\multirow{1}{*}{ \textbf{\underline{ 33.57\%}}}} & \multicolumn{1}{c}{ \textbf{\underline{ 4.508}} } \\
			&  	\multicolumn{1}{l}{\multirow{1}{*}{\parbox{4cm}{\centering TFM + Softplus intensity \cite{zuo2020transformer}}}} & \multicolumn{1}{c}{\multirow{1}{*}{ 33.27\%}} & \multicolumn{1}{c}{  35.01 } \\
			
			\midrule[1pt]
			
			\multicolumn{1}{l}{\multirow{5}{*}{\parbox{1.5cm}{\centering ATM } }} &	\multicolumn{1}{l}{\multirow{1}{*}{\parbox{4cm}{\centering \textbf{LSTM + Exp interarrival}}}} & \multicolumn{1}{c}{ \multirow{1}{*}{  \textbf{\underline{ 92.51\%}}}}  &    \multicolumn{1}{c}{ \underline{\textbf{3.105}} } \\ 
			&	\multicolumn{1}{l}{\multirow{1}{*}{\parbox{4cm}{\centering RNN + Exp intensity \cite{du2016recurrent}} }} & \multicolumn{1}{c}{\multirow{1}{*}{ 76.64\%}  }   &    \multicolumn{1}{c}{ 7.150}
			\\
			&	\multicolumn{1}{l}{\multirow{1}{*}{\parbox{4cm}{\centering LSTM + Sigmoid intensity \cite{mei2017neural}} }}  &\multicolumn{1}{c}{\multirow{1}{*}{ 33.78\%}  }  &   \multicolumn{1}{c}{ 7.558}\\\cmidrule{2-4}
			&  	\multicolumn{1}{l}{\multirow{1}{*}{\parbox{4cm}{\centering \textbf{TFM + Exp interarrival} }}} &  \multicolumn{1}{c}{\multirow{1}{*}{ \textbf{\underline{ 90.60\%}}}} & \multicolumn{1}{c}{ \textbf{\underline{ 3.245}} } \\
			&  	\multicolumn{1}{l}{\multirow{1}{*}{\parbox{4cm}{\centering TFM + Softplus intensity \cite{zuo2020transformer}}}} & \multicolumn{1}{c}{\multirow{1}{*}{ 68.76\%}} & \multicolumn{1}{c}{  4.534 } \\
			
			\midrule[1pt]
			
			\multicolumn{1}{l}{\multirow{5}{*}{\parbox{1.5cm}{\centering IPTV } }} &	\multicolumn{1}{l}{\multirow{1}{*}{\parbox{4cm}{\centering \textbf{LSTM + Exp interarrival}}}} & \multicolumn{1}{c}{ \multirow{1}{*}{  \textbf{\underline{ 76.20\%}}}}  &    \multicolumn{1}{c}{ \underline{\textbf{11.238}} } \\ 
			&	\multicolumn{1}{l}{\multirow{1}{*}{\parbox{4cm}{\centering RNN + Exp intensity \cite{du2016recurrent}} }} & \multicolumn{1}{c}{\multirow{1}{*}{ 57.57\%}  }   &    \multicolumn{1}{c}{ 34.382}
			\\
			&	\multicolumn{1}{l}{\multirow{1}{*}{\parbox{4cm}{\centering LSTM +  Sigmoid intensity \cite{mei2017neural}} }}  &\multicolumn{1}{c}{\multirow{1}{*}{ 31.05\%}  }  &   \multicolumn{1}{c}{ 19.929}\\\cmidrule{2-4}
			&  	\multicolumn{1}{l}{\multirow{1}{*}{\parbox{4cm}{\centering \textbf{TFM + Exp interarrival} }}} &  \multicolumn{1}{c}{\multirow{1}{*}{ \textbf{\underline{ 76.20\%}}}} & \multicolumn{1}{c}{ \textbf{\underline{ 10.188}} } \\
			&  	\multicolumn{1}{l}{\multirow{1}{*}{\parbox{4cm}{\centering TFM + Softplus intensity \cite{zuo2020transformer}}}} & \multicolumn{1}{c}{\multirow{1}{*}{ 71.94\%}} & \multicolumn{1}{c}{  31.325 } \\
			
			\midrule[1pt]
			
			\multicolumn{1}{l}{\multirow{5}{*}{\parbox{1.5cm}{\centering Weeplace } }} &	\multicolumn{1}{l}{\multirow{1}{*}{\parbox{4cm}{\centering \textbf{LSTM + Exp interarrival}}}} & \multicolumn{1}{c}{ \multirow{1}{*}{  \textbf{\underline{ 31.86\%}}}}  &    \multicolumn{1}{c}{ \underline{\textbf{6.777}} } \\ 
			&	\multicolumn{1}{l}{\multirow{1}{*}{\parbox{4cm}{\centering RNN + Exp intensity \cite{du2016recurrent}} }} & \multicolumn{1}{c}{\multirow{1}{*}{ 22.07\%}  }   &    \multicolumn{1}{c}{ 7.162}
			\\
			&	\multicolumn{1}{l}{\multirow{1}{*}{\parbox{4cm}{\centering LSTM + Sigmoid intensity \cite{mei2017neural}} }}  &\multicolumn{1}{c}{\multirow{1}{*}{ 25.71\%}  }  &   \multicolumn{1}{c}{ 6.773}\\\cmidrule{2-4}
			&  	\multicolumn{1}{l}{\multirow{1}{*}{\parbox{4cm}{\centering \textbf{TFM + Exp interarrival} }}} &  \multicolumn{1}{c}{\multirow{1}{*}{ \textbf{\underline{ 32.62\%}}}} & \multicolumn{1}{c}{ \textbf{\underline{ 6.571}} } \\
			&  	\multicolumn{1}{l}{\multirow{1}{*}{\parbox{4cm}{\centering TFM + Softplus intensity  \cite{zuo2020transformer}}}} & \multicolumn{1}{c}{\multirow{1}{*}{ 29.24\%}} & \multicolumn{1}{c}{  51.78 } \\

			\bottomrule[2pt]
	\end{tabular}}
	\label{loss_type}
\end{table}

	\section{Conclusion}
	In this paper, we describe an interesting performance saturation phenomenon when training neural marked point process models: performance often becomes stagnated at some point, and cannot be improved any more by making the network more complicated. From the generalization error analysis and experimental results, we conclude our paper with two suggestions for using neural marked point process models: first, for some cases, a simple network structure can perform as well as complicated ones, but more efficiently; second, using a proper probabilistic assumption is as equally, if not more, important as improving the network structure. In the future, we would like to investigate the reason of this phenomenon theoretically.
	
	\section*{Acknowledgment}
	This research/project is supported by: (1) the National Research Foundation, Singapore under its Industry Alignment Fund – Pre-positioning (IAF-PP) Funding Initiative; (2) NTU Singapore Nanyang Assistant Professorship (NAP) grant M4081532.020; (3) Singapore MOE AcRF Tier-1 grant 2018-T1-002-143 (RG131/18 (S)); and (4) Alibaba-NTU Singapore Joint Research Institute, Nanyang Technological University. Any opinions, findings and conclusions or recommendations expressed in this material are those of the author(s) and do not reflect the views of National Research Foundation, Singapore. 
	
\balance	
\vfill\eject
\bibliographystyle{ACM-Reference-Format}
\bibliography{cpp}

\end{document}